\documentclass[conference]{IEEEtran}
\IEEEoverridecommandlockouts

\usepackage{cite}
\usepackage{amsmath,amssymb,amsfonts}
\usepackage{algorithmic}
\usepackage{graphicx}
\usepackage{textcomp}
\usepackage{xcolor}
\def\BibTeX{{\rm B\kern-.05em{\sc i\kern-.025em b}\kern-.08em
    T\kern-.1667em\lower.7ex\hbox{E}\kern-.125emX}}

\usepackage[caption=false]{subfig}
\usepackage{textcomp}
\usepackage{stfloats}
\usepackage{url}
\usepackage{verbatim}
\usepackage{graphicx}
\usepackage{cite}

\usepackage[english]{babel}
\usepackage{soul,color}
\soulregister\cite7
\soulregister\ref7
\soulregister\ac7
\soulregister\acp7
\soulregister\underline7

\usepackage{threeparttable}
\usepackage{url}
\usepackage{cite}
\DeclareMathAlphabet{\pazocal}{OMS}{zplm}{m}{n}
\usepackage{nicefrac}
\usepackage{optidef}
\usepackage{gensymb}
\usepackage{xurl}
\usepackage{booktabs}
\usepackage{xcolor}

\begin{document}

\title{Bypassing the CSI Bottleneck: MARL-Driven Spatial Control for Reflector Arrays\\
\thanks{This material is based upon work supported by the U.S. Department of Energy, Office of Science, Office of Advanced Scientific Computing Research, Early Career Research Program under Award Number DE-SC-0023957.}
}


\author{\IEEEauthorblockN{Hieu Le}
\IEEEauthorblockA{\textit{Electrical and Computer Engineering} \\
\textit{Texas A\&M University}\\
College Station, Texas, USA \\
hieult@tamu.edu} \\

\\
\IEEEauthorblockN{Jian Tao}
\IEEEauthorblockA{\textit{School of Performance, } \\
\textit{Visualization, and Fine Arts} \\
\textit{Texas A\&M University}\\
College Station, Texas, USA \\
jtao@tamu.edu }
\and
\IEEEauthorblockN{Mostafa Ibrahim}
\IEEEauthorblockA{\textit{Engineering Technology } \\
\textit{and Industrial Distribution} \\
\textit{Texas A\&M University}\\
College Station, Texas, USA \\
mostafa.ibrahim@tamu.edu }
\\
\IEEEauthorblockN{Sabit Ekin}
\IEEEauthorblockA{\textit{Engineering Technology, and} \\
\textit{Electrical and Computer Engineering} \\
\textit{Texas A\&M University}\\
College Station, Texas, USA \\
sabitekin@tamu.edu }
\and
\IEEEauthorblockN{Oguz Bedir}
\IEEEauthorblockA{\textit{Electrical and Computer Engineering} \\
\textit{Texas A\&M University}\\
College Station, Texas, USA \\
oguzbedir@tamu.edu }
}

\maketitle

\begin{abstract}
Reconfigurable Intelligent Surfaces (RIS) are pivotal for next-generation smart radio environments, yet their practical deployment is severely bottlenecked by the intractable computational overhead of Channel State Information (CSI) estimation. To bypass this fundamental physical-layer barrier, we propose an AI-native, data-driven paradigm that replaces complex channel modeling with spatial intelligence. This paper presents a fully autonomous Multi-Agent Reinforcement Learning (MARL) framework to control mechanically adjustable metallic reflector arrays. By mapping high-dimensional mechanical constraints to a reduced-order virtual focal point space, we deploy a Centralized Training with Decentralized Execution (CTDE) architecture. Using Multi-Agent Proximal Policy Optimization (MAPPO), our decentralized agents learn cooperative beam-focusing strategies relying on user coordinates, achieving CSI-free operation. High-fidelity ray-tracing simulations in dynamic non-line-of-sight (NLOS) environments demonstrate that this multi-agent approach rapidly adapts to user mobility, yielding up to a 26.86 dB enhancement over static flat reflectors and outperforming single-agent and hardware-constrained DRL baselines in both spatial selectivity and temporal stability. Crucially, the learned policies exhibit good deployment resilience, sustaining stable signal coverage even under 1.0-meter localization noise. These results validate the efficacy of MARL-driven spatial abstractions as a scalable, highly practical pathway toward AI-empowered wireless networks.
\end{abstract}


\section{Introduction}

The exponential growth in wireless data traffic has pushed traditional communication systems to their fundamental limits, largely because conventional approaches treat the wireless propagation environment as a static, uncontrollable obstacle \cite{direnzo:2020}. Reconfigurable intelligent surfaces (RIS) represent a paradigm shift toward programmable wireless environments, where passive objects become active participants in signal propagation \cite{bjornson2022reconfigurable}. However, despite their theoretical promise, practical RIS implementations face critical deployment barriers. The most significant challenge is the overwhelming complexity of channel state information (CSI) estimation, which requires precise electromagnetic characterization across hundreds of reflecting elements and creates a computational overhead that scales exponentially \cite{pan2022overview}. Furthermore, systems relying on constructive interference demand highly sophisticated hardware, including high-resolution phase shifters and strict synchronization mechanisms \cite{kim2022practical}. 

Multiple sophisticated approaches have attempted to address this CSI bottleneck. Typical cascaded channel estimation methods suffer from pilot overheads that reach hundreds to thousands of symbols, creating prohibitive spectral efficiency losses \cite{a9400843}. More recently, Deep Reinforcement Learning (DRL) has emerged to optimize RIS phase shifts and circumvent traditional limitations \cite{huang2020reconfigurable}. Yet, even these advanced DRL implementations fundamentally rely on partial channel knowledge at the reflecting surface \cite{choi2024deep} or require extensive offline training data \cite{sheen2021deep}, increasing system complexity and energy consumption.

To bypass these fundamental barriers, this work introduces a different approach that eliminates CSI estimation entirely by operating at a higher abstraction level. As comprehensively detailed in the work \cite{aa11322690}, our method employs mechanically adjustable metallic reflectors \cite{a8972365, le2024guiding} that dynamically modify their orientation. By leveraging user location information to optimize reflection geometries, this architecture provides inherent broadband operation and simplifies control mechanisms using commercial off-the-shelf servo systems. User location information can be obtained by using modern localization techniques such as Ultra-Wide Band (UWB) Technology, which can achieve sub-meter localization accuracy.

We formulate this large-scale propagation control as a cooperative multi-agent reinforcement learning (MARL) problem \cite{busoniu2008comprehensive, zhang2018fully}. By utilizing a centralized training with decentralized execution (CTDE) paradigm, individual agents control specific reflector segments and autonomously learn to cooperate. This approach decomposes the complex system-wide optimization into manageable sub-problems, enabling scalable and adaptive operation without requiring explicit coordination protocols or information exchange between elements \cite{a10261304, a10654286}. Additionally, practical deployment feasibility is enhanced by the widespread availability of neural network processing hardware and specialized deep learning accelerators in modern systems \cite{nasari2022benchmarking, le2024insight, qualcomm2024unlocking}.

Our primary contributions are summarized as follows:
\begin{itemize}
    \item \textit{CSI-free spatial control:} We formulate reflector control as a multi-agent Markov decision process (MA-MDP), enabling DRL techniques to control wireless propagation through spatial intelligence rather than complex electromagnetic channel estimation.
    \item \textit{Signal enhancement:} Through extensive high-fidelity ray-tracing simulations, our multi-agent beam-focusing framework demonstrates up to a $26.86\,\mathrm{dB}$ received signal strength indicator (RSSI) improvement over static flat reflectors. Furthermore, the proposed method outperforms alternative single-agent and hardware-constrained multi-agent DRL baselines, proving that decentralized spatial task decomposition is essential for maximizing RSSI.
    \item \textit{Practical robustness:} We validate the system's deployment viability by demonstrating significant resilience to localization uncertainties and dynamic user mobility, confirming performance advantages even with localization accuracy limitations.
\end{itemize}

\section{System Model and Problem Formulation}
\label{sec:sys_model}

\subsection{System Architecture and Focal Point Control}
\label{subsec:sys_arch}

\begin{figure}[!ht]
    \includegraphics[width=0.95\linewidth]{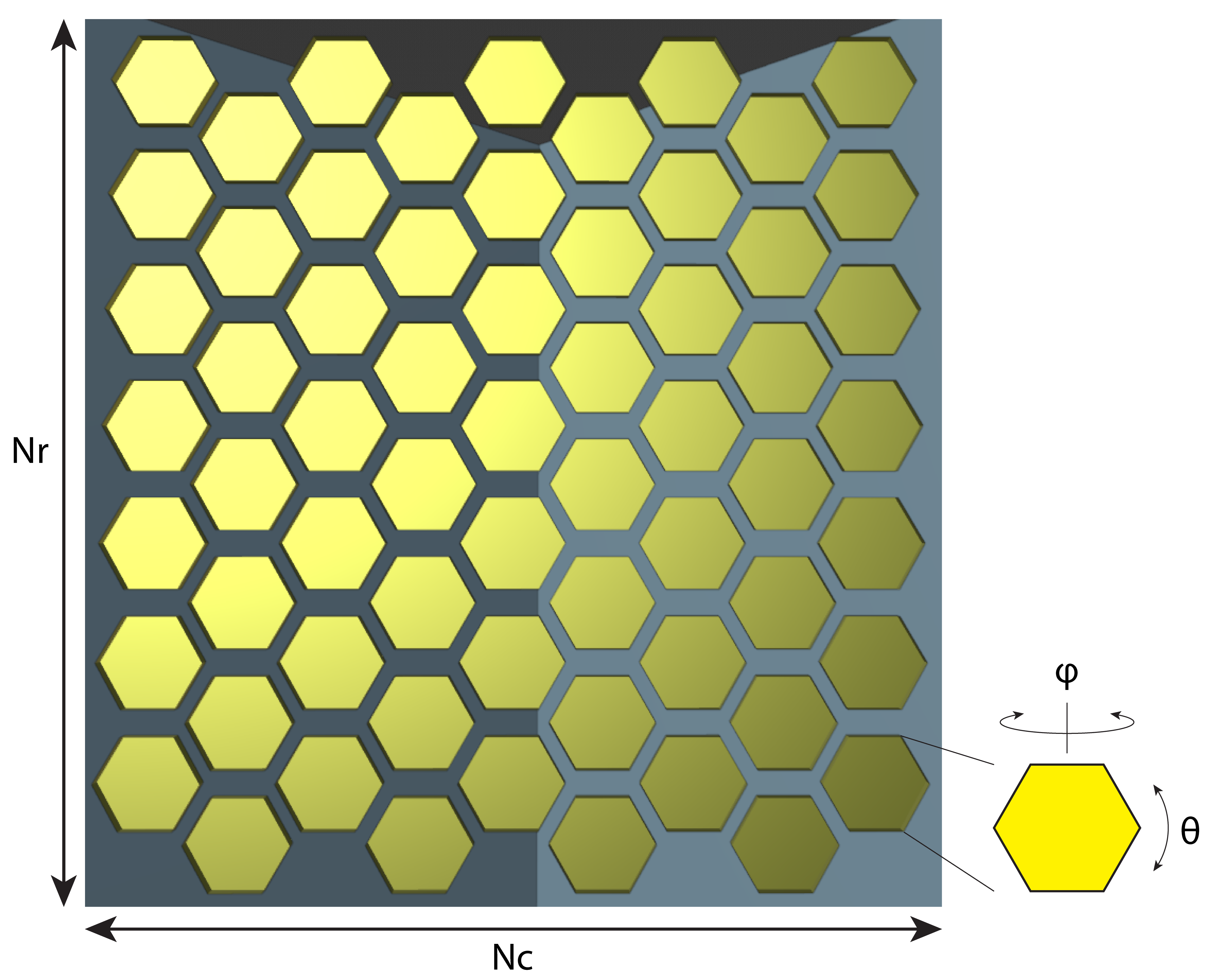}
    \caption{Reflector Design.}
    \label{Figure:reflector}
\end{figure}

We consider reflector arrays composed of an $N_r \times N_c$ array of hexagonal metallic tiles (Fig.~\ref{Figure:reflector}). Unlike conventional electronic systems, these tiles are mechanically adjustable in both elevation and azimuth to physically steer millimeter-wave (mmWave) beams. This mechanical reconfiguration enables precise control over electromagnetic wavefront manipulation, completely bypassing the need for complex RF circuits and electronic phase shifters.

A fundamental challenge in scaling mechanically adjustable arrays is the overwhelming dimensionality of the control space. To address this, we introduce a focal point control mechanism that abstracts the complex, per-tile angular optimization into a simplified geometric framework. Rather than individually tuning the rotation of each element, the reflector array is partitioned into $L$ disjoint segments, where an intelligent agent $l$ is assigned to control a virtual, movable focal point $\mathbf{f}_{l,t}$ in three-dimensional space.

For reflector element $(i,j)$ belonging to agent $l$, elevation $\theta_{i,j,t}$ and azimuth $\phi_{i,j,t}$ can be calculated using a bisector vector as:

\begin{equation}
    \overrightarrow{n_{i,j,t}} = \frac{1}{2} \left( 
    \frac{\overrightarrow{s} - \overrightarrow{ r_{i,j}}}{\|s - r_{i,j}\|} + \frac{\overrightarrow{f_{l,t}} - \overrightarrow{ r_{i,j}}}{\|f_{l,t} - r_{i,j}\|} \right),
    \label{eq:normal}
\end{equation}

\begin{equation}
\label{eq:phi_cal}
\phi_{i,j,t} = \operatorname{atan2}\big( \overrightarrow{n_{i,j,t}} \cdot \hat{y}, \overrightarrow{n_{i,j,t}} \cdot \hat{x}\big),
\end{equation}
\begin{equation}
\label{eq:theta_cal}
\theta_{i,j,t} = \arccos\!\left( \overrightarrow{n_{i,j,t}} \cdot \hat{z} \right),
\end{equation}
where $r_{i,j} \in \mathbb{R}^3$ is the position of tile $(i,j)$, $s \in \mathbb{R}^3$ denotes the access point location, and $\hat{x}, \hat{y}, \hat{z}$ are unit vectors defining the coordinate frame.

Physical constraints impose:
\begin{equation}
\phi_{i,j,t} \in [\phi_{\min}, \phi_{\max}], \quad
\theta_{i,j,t} \in [\theta_{\min}, \theta_{\max}], \quad
 \forall i,j.
\end{equation}

The spatial evolution of this focal point dynamically updates according to the agent's action $\mathbf{a}_{l,t}$, expressed as
\begin{equation}
    \mathbf{f}_{l,t+1} = \mathbf{f}_{l,t} + \mathbf{a}_{l,t}.
\end{equation}

Based on the coordinates of $\mathbf{f}_{l,t}$, the required elevation and azimuth angles for all tiles within segment $l$ are deterministically computed using standard reflection geometry. This spatial abstraction yields a high dimensionality reduction: the optimization space is compressed from $2 N_r N_c$ distinct angular parameters down to merely $3L$ focal point coordinates. Because $L \ll N_r N_c$ in practical deployments, this geometrical abstraction accelerates the convergence of the subsequent learning framework while natively enforcing the mechanical limitations of the servo actuators.

\begin{figure}[!t]
    \centering
    \includegraphics[width=1\linewidth]{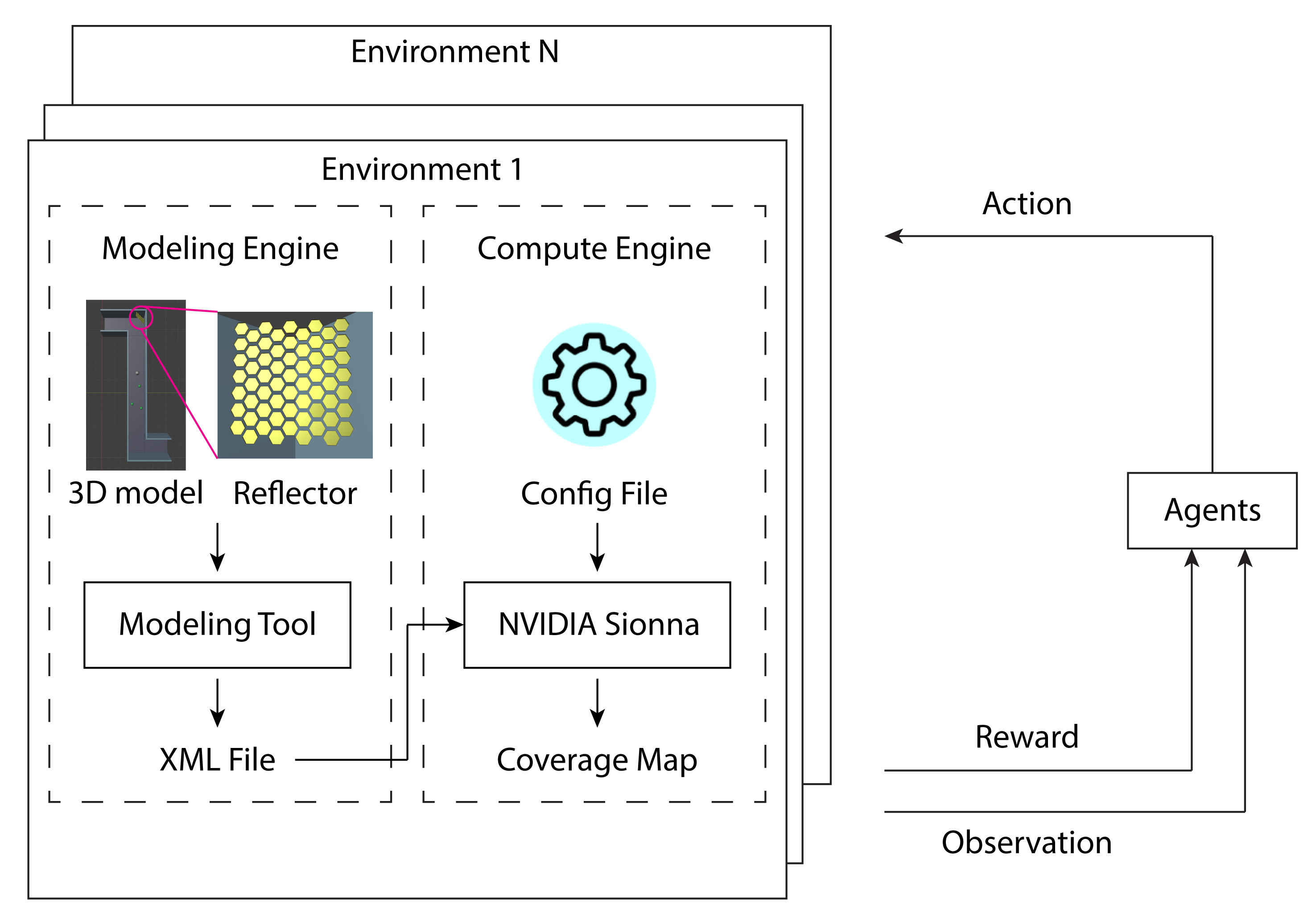}
    \caption{Workflow of the Deep Reinforcement Learning and the integrated environment with Sionna and Blender.}
    \label{Figure:workflow}
\end{figure}

\subsection{Multi-Agent MDP Formulation}

To optimize the reflector array dynamically without relying on explicit, high-overhead CSI estimation, we formulate the propagation control problem as a cooperative MA-MDP 
\((\mathcal{S}, \{\mathcal{O}_l\}_{l=1}^L, \{\mathcal{A}_l\}_{l=1}^L, P, R, \gamma)\),
where $\mathcal{S}$ is the global state space, $\mathcal{O}_l$ and $\mathcal{A}_l$ denote the local observation and action spaces of agent $l$, $P$ is the state transition function, $R$ is the shared reward function, and $\gamma \in (0,1)$ is the discount factor.
The environment comprises a set of $L$ intelligent agents, each governing a specific segment of the reflector surface.


\subsubsection{State and Local Observations}
At each time step $t$, the global state $s_t \in \mathcal{S}$ captures the complete spatial configuration of the environment, defined as the three-dimensional positions of all $K$ users, the fixed coordinates of the reflector segments, and the current spatial focal points of all $L$ agents. To ensure the system remains scalable and deployable without prohibitive inter-agent communication overhead, each agent $l$ executes its policy based strictly on a localized observation $o_{l,t}$. This observation is restricted to the position of its assigned user, its own reflector segment location, and its current focal point.

\subsubsection{Action Space}
Operating within the spatial abstraction, the continuous action space $\mathcal{A}_l$ for agent $l$ is defined purely as the 3D displacement vector applied to its focal point:
\begin{equation}
    \mathbf{a}_{l,t} = [\Delta f_{l,x,t}, \Delta f_{l,y,t}, \Delta f_{l,z,t}]^T
\end{equation}
bounded by a maximum allowable displacement $\delta_{\max}$. Because the mechanical limitations of the servo motors (e.g., maximum rotation angles) are handled geometrically by bounding the allowable focal point region, the agent's action space is naturally constrained to the physically feasible set. This eliminates the need for complex penalty terms typically required in constrained MDPs, stabilizing the learning process.

\subsubsection{Reward Function}
The cooperative goal of the multi-agent system is to maximize the aggregate signal quality while ensuring targeted coverage for individual users. To achieve this, the reflector array is partitioned such that each agent $l$ is explicitly pre-assigned to focus on a specific user, denoted as $k(l)$. Therefore, rather than a strictly common global reward, each agent receives a hybrid reward signal. This reward combines the global average RSSI across all $K$ users with a local term representing the specific RSSI of its assigned user $k(l)$:
\begin{equation}
    R_k(s_t, \mathbf{a}_t) = \frac{1}{K} \sum_{k'=1}^{K} P_{r,k'}(s_t, \mathbf{a}_t) + P_{r,k}(s_t, \mathbf{a}_t),
\end{equation}
where $P_{r,k}(s_t, \mathbf{a}_t)$ denotes the RSSI achieved for user $k$ under the joint focal point configuration. By maximizing this expected discounted cumulative hybrid reward, the agents implicitly learn to balance explicitly serving their assigned targets with cooperating to exploit system-wide indirect reflections.

\subsection{Multi-Agent Reinforcement Learning Framework}

To optimize the joint focal point configuration, we employ multi-agent proximal policy optimization (MAPPO \cite{yu2022surprising}) within a CTDE paradigm. This architecture alleviates the non-stationarity problem inherent in multi-agent environments, where simultaneous learning by multiple agents causes the environment dynamics to appear unstable from any single agent's perspective.

\subsubsection{Centralized Training with Decentralized Execution}
During the offline training phase, the CTDE framework utilizes a centralized critic network that has access to the complete global state $s_{\text{global}}$. By evaluating the joint actions of all agents against the global state, the centralized critic computes 
accurate value estimates and stable advantage functions. However, during online deployment (execution), the system transitions to a fully decentralized model, where each agent $l$ independently executes its learned policy $\pi_{\theta_l}(a_l \mid o_l)$ relying solely on its local observation $o_l$. This ensures the system remains practically deployable and scalable, preserving the sophisticated coordination learned during training without requiring any explicit inter-agent communication or centralized control overhead.

\subsubsection{Policy Optimization}
The specific learning algorithm, MAPPO \cite{yu2022surprising}, extends the stability of single-agent PPO to the multi-agent domain. A key feature of MAPPO is its use of a clipped surrogate objective function, which bounds the policy update ratio and thus the magnitude of policy changes at each training step. In a cooperative array where the optimal orientation of one reflector segment heavily depends on the orientations of the others, this clipping mechanism is crucial, as it prevents large, destructive policy updates that could destabilize the emergent coordination patterns between the reflecting elements.

\section{Results and Discussion}
\label{sec:results_and_discussion}

\subsection{Simulation Setup and Learning Progression}
To evaluate the proposed MARL framework, we model a $60\,\mathrm{GHz}$ mmWave downlink, with a AP having a transmit power of 5 mW, in an L-shaped hallway under non-line-of-sight (NLOS) conditions as shown in Fig.~\ref{Figure:workflow}. The environment features a single access point, three mobile users, and a 72-tile hexagonal reflector array. Signal propagation and multipath interactions are simulated using the NVIDIA Sionna ray-tracing engine, incorporating realistic material properties (plasterboard, concrete, and wood) based on ITU recommendations.

The MAPPO framework is implemented using fully connected neural networks with two hidden layers of 256 neurons and ReLU activations for both the actor and critic architectures. The networks are trained using the Adam optimizer with a constant learning rate of $2.0 \times 10^{-4}$ and a minibatch size of 200 sampled from a replay buffer of 1000 transitions. To ensure stable multi-agent coordination and prevent destructive policy updates, the PPO clipping parameter is set to $\epsilon_{CLIP} = 0.2$, alongside a discount factor of $\gamma = 0.985$, a GAE parameter of $\lambda_{GAE} = 0.9$, and value and entropy coefficients of 0.5 and $1.0 \times 10^{-4}$, respectively.

\begin{table}[ht!]
\caption{MAPPO Hyperparameters.}
\centering
\begin{tabular}{l|l}
    \hline
    Parameter & Value \\
    \hline
    Number of neurons (hidden layers) & 256 \\
    Optimizer & Adam \\
    Learning rate & $2.0 \times 10^{-4}$ \\
    Learning rate schedule & Constant \\
    Discount factor ($\gamma$) & 0.985 \\
    Replay buffer size & 1000 \\
    Minibatch size & 200 \\
    Activation function & ReLU \\
    PPO clipping parameter ($\epsilon_{CLIP}$) & 0.2 \\
    GAE parameter ($\lambda_{GAE}$) & 0.9 \\
    Value function coefficient ($c_1$) & 0.5 \\
    Entropy coefficient ($c_2$) & $1.0 \times 10^{-4}$ \\
    \hline
\end{tabular}
\label{table:hyperparameters}
\end{table}


We assess the learning efficiency of our proposed multi-agent beam-focusing (\textit{beam-focusing-ma}) approach against two baselines: a single-agent beam-focusing (\textit{beam-focusing-sa}) framework and a constrained multi-agent column-based (\textit{column-based-ma}) approach. The two beam-focusing configurations enable full spatial degrees of freedom, allowing independent elevation and azimuth adjustments for every individual tile. Conversely, the \textit{column-based-ma} method models a practical cost-performance trade-off by restricting azimuth rotation to column-level control. In this constrained configuration, all reflecting elements within a given vertical column strictly share a single, uniform azimuth angle, requiring only one azimuth servo per column, while preserving independent elevation control for each individual tile. This design significantly reduces the mechanical complexity and control dimensionality of the array, albeit at the expense of finer spatial granularity.

\begin{figure}[!t]
    \centering
    \captionsetup{justification=centering}
    \includegraphics[width=1\linewidth]{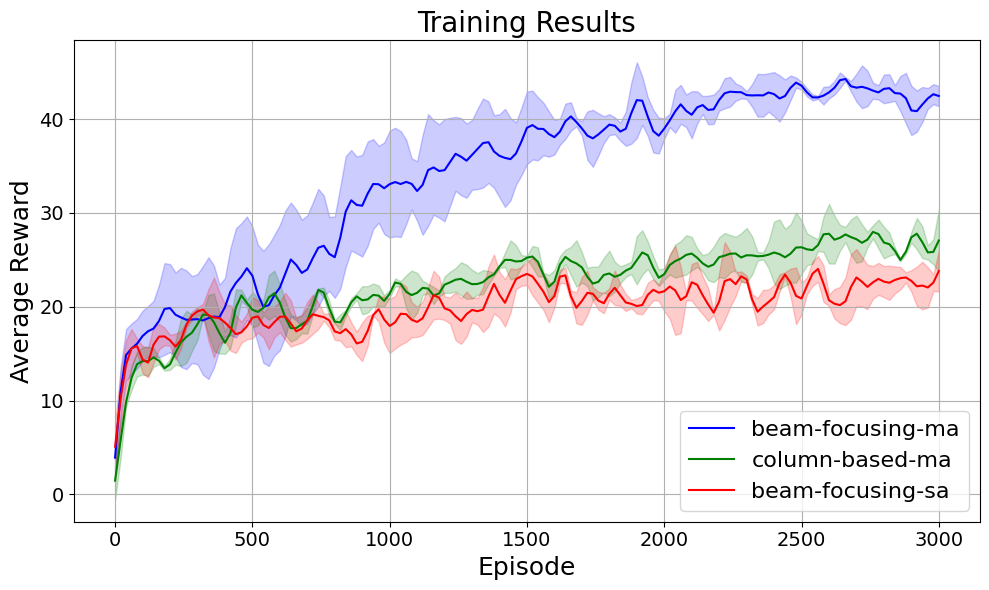}
    \caption{{Reward convergence for proposed and baseline algorithms for reflecting element control. Our proposed beam-focusing-ma framework achieves a peak convergence value of 42, outperforming the beam-focusing-sa and column-based-ma baseline methods (both converging near 25). Mean cumulative rewards are depicted by solid lines, with shaded envelopes representing the variance across multiple training trials.
}}
    \label{Figure:training}
\end{figure}

As illustrated in the training progression over 3000 episodes (Fig.~\ref{Figure:training}), the beam-focusing-ma approach demonstrates superior convergence characteristics. The multi-agent formulation quickly increases during the first 1000 episodes and converges to an average cumulative reward of approximately 42. In stark contrast, both the hardware-constrained \textit{column-based-ma} and the \textit{beam-focusing-sa} baselines plateau earlier, converging around reward values of 27 and 24, respectively. The substantial performance gap between the multi-agent and single-agent beam-focusing models underscores the inherent difficulty of centralizing complex, high-dimensional control tasks. By decomposing the optimization problem into agent-specific sub-tasks through the CTDE paradigm, our framework successfully mitigates the dimensionality curse, allowing individual agents to efficiently learn specialized coordination patterns to serve their assigned users.

\begin{figure}[!t]
\centering
\subfloat[No Reflector.\protect\\(RSSI: $-110.50$ dBm)]
{\includegraphics[width=0.3\linewidth]{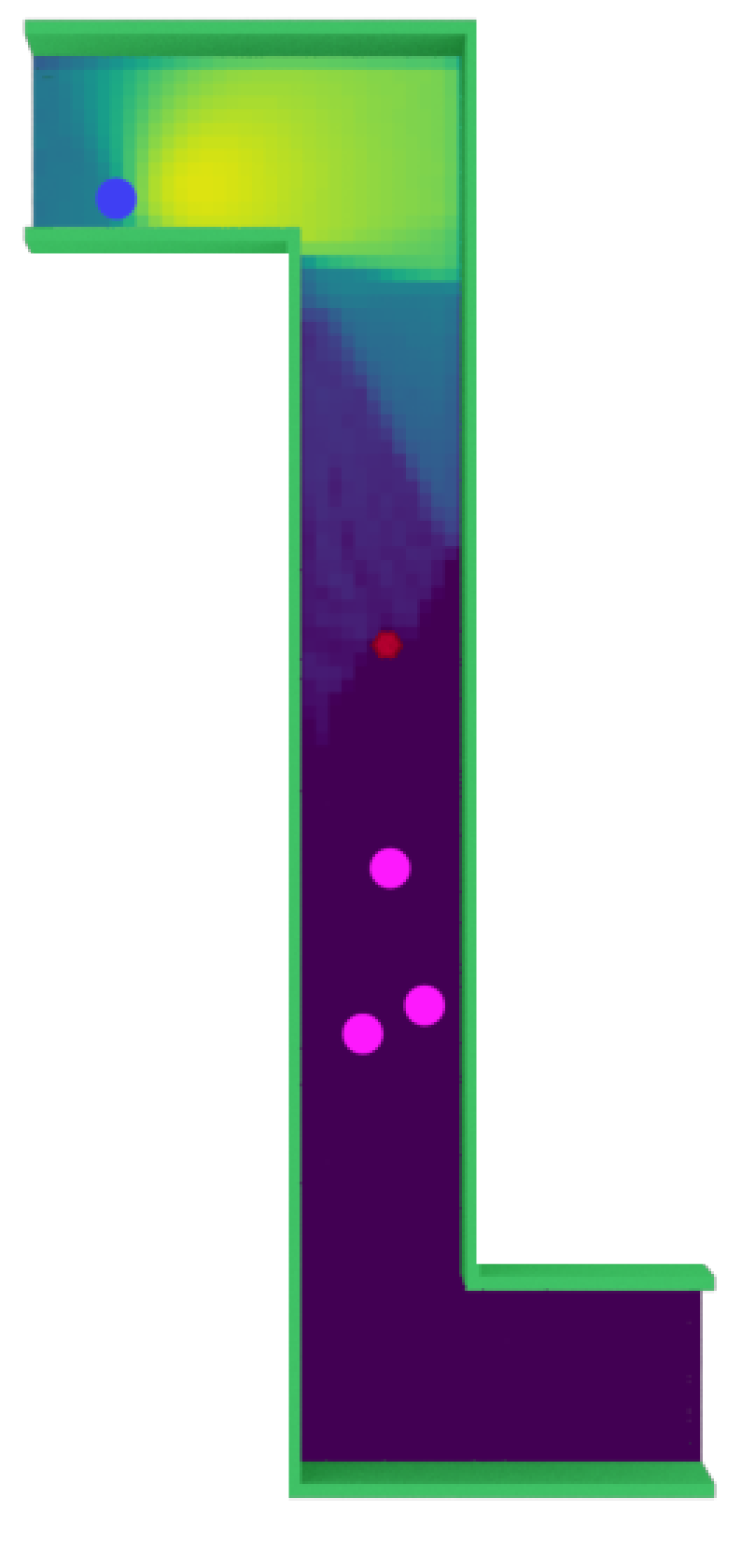}%
\label{Figure:no_reflector}}
\hfill
\subfloat[Flat Reflector.\protect\\(RSSI: $-94.40$ dBm)]
{\includegraphics[width=0.3\linewidth]{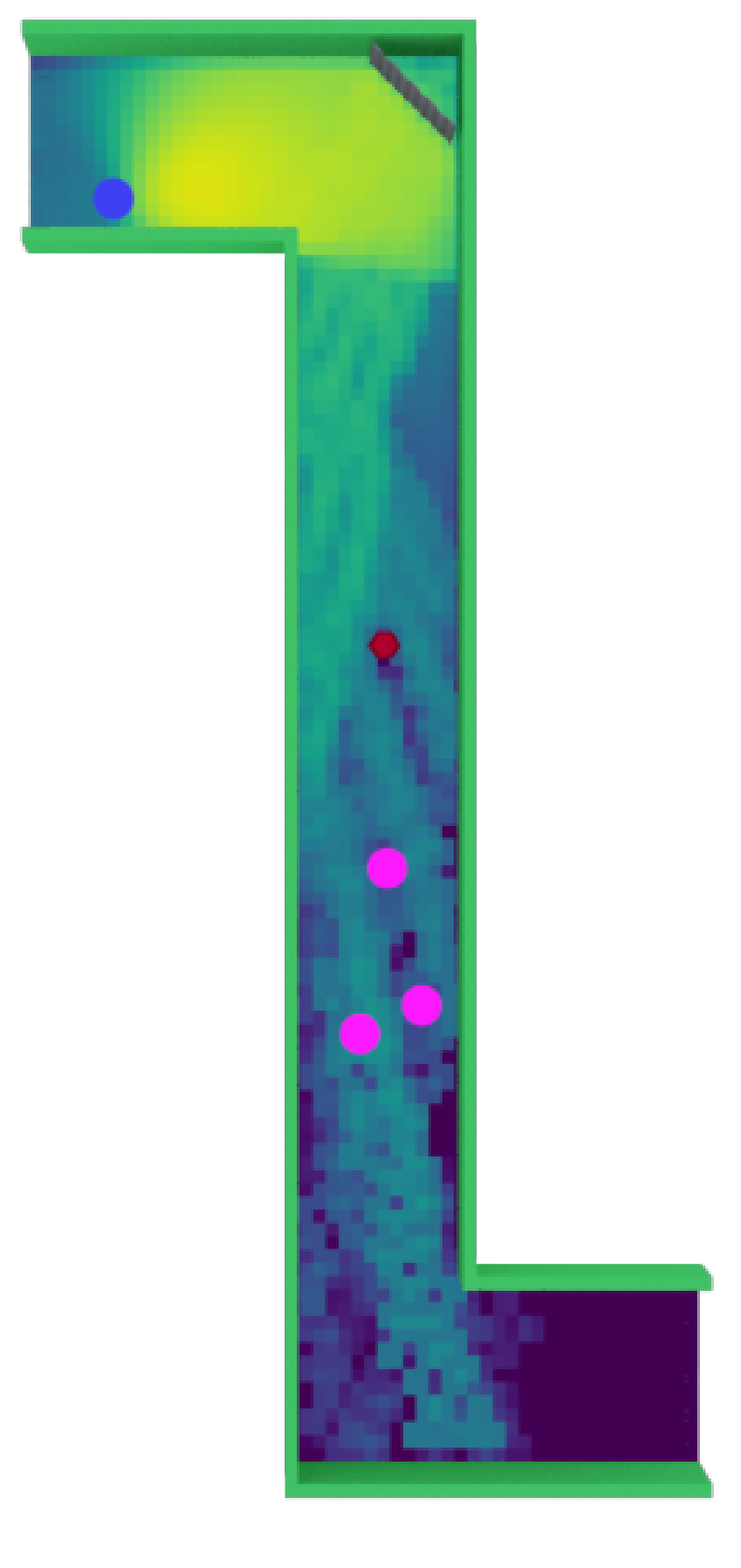}%
\label{Figure:flat_reflector}}
\hfill
\subfloat[SA Focusing.\protect\\(RSSI: $-72.48$ dBm)]
{\includegraphics[width=0.3\linewidth]{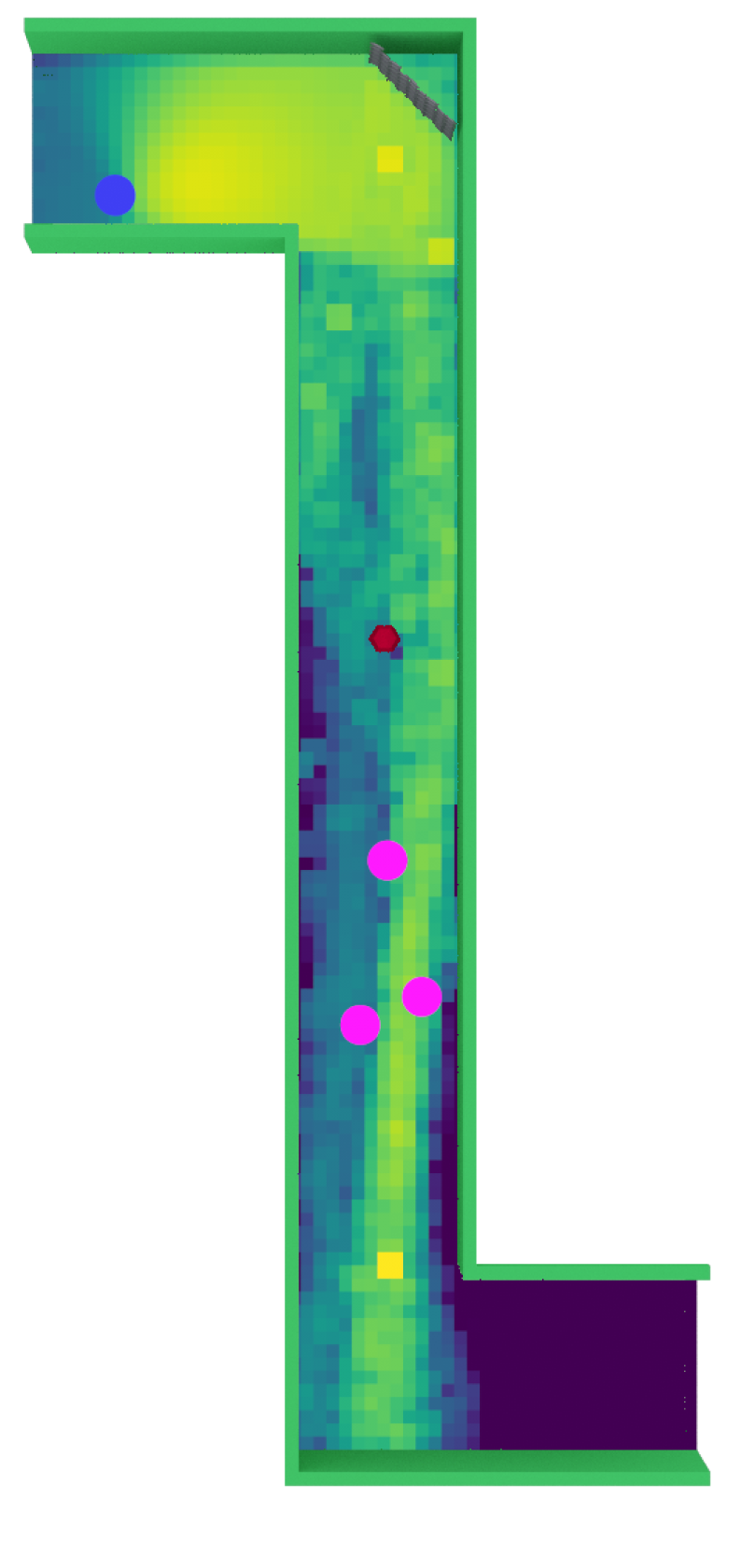}%
\label{Figure:beamfocusing_sa}}

\subfloat[MA Col.\protect\\(RSSI: $-74.79$ dBm)]
{\includegraphics[width=0.3\linewidth]{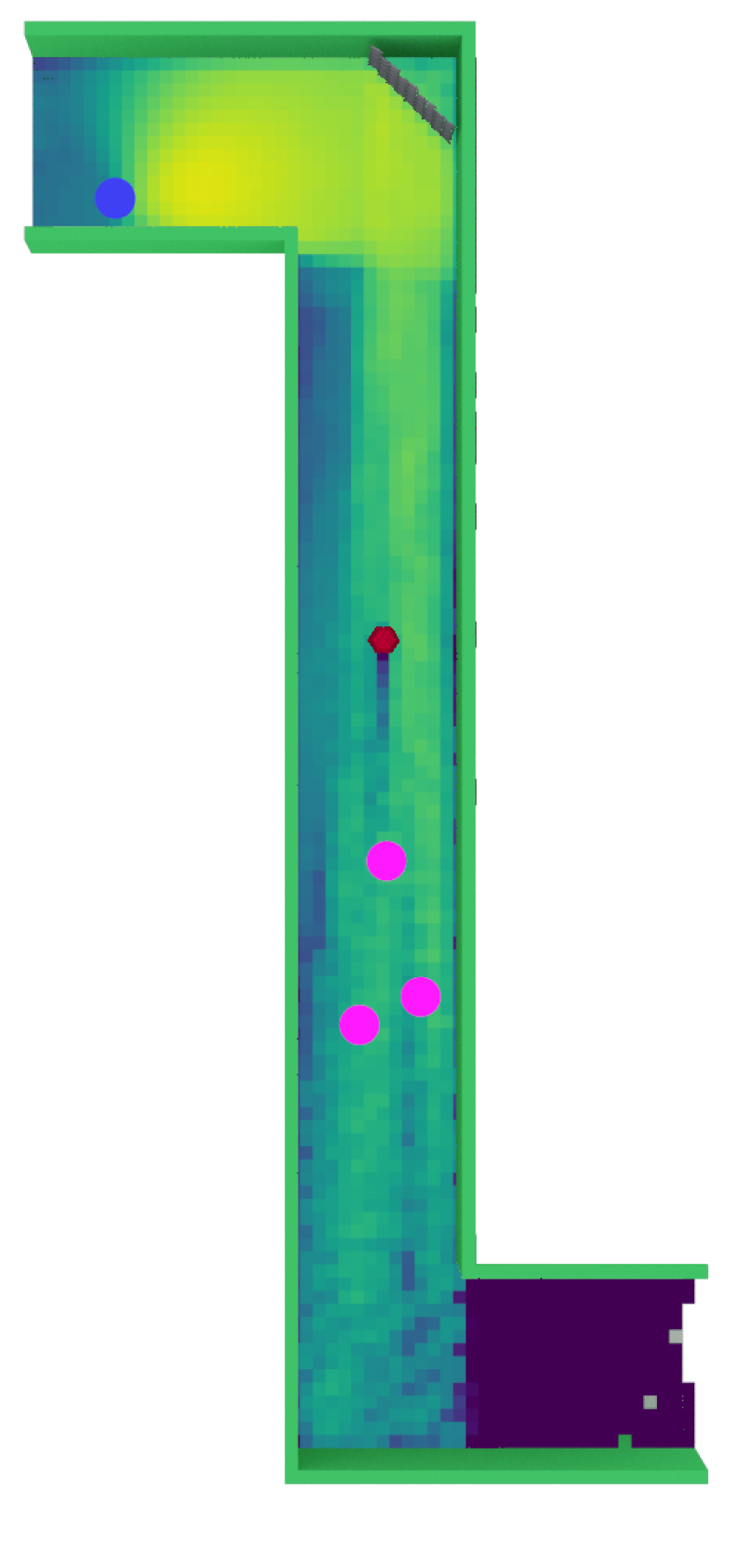}%
\label{Figure:col_ma}}
\hfill
\subfloat[MA Focusing.\protect\\(RSSI: $-67.54$ dBm)]
{\includegraphics[width=0.3\linewidth]{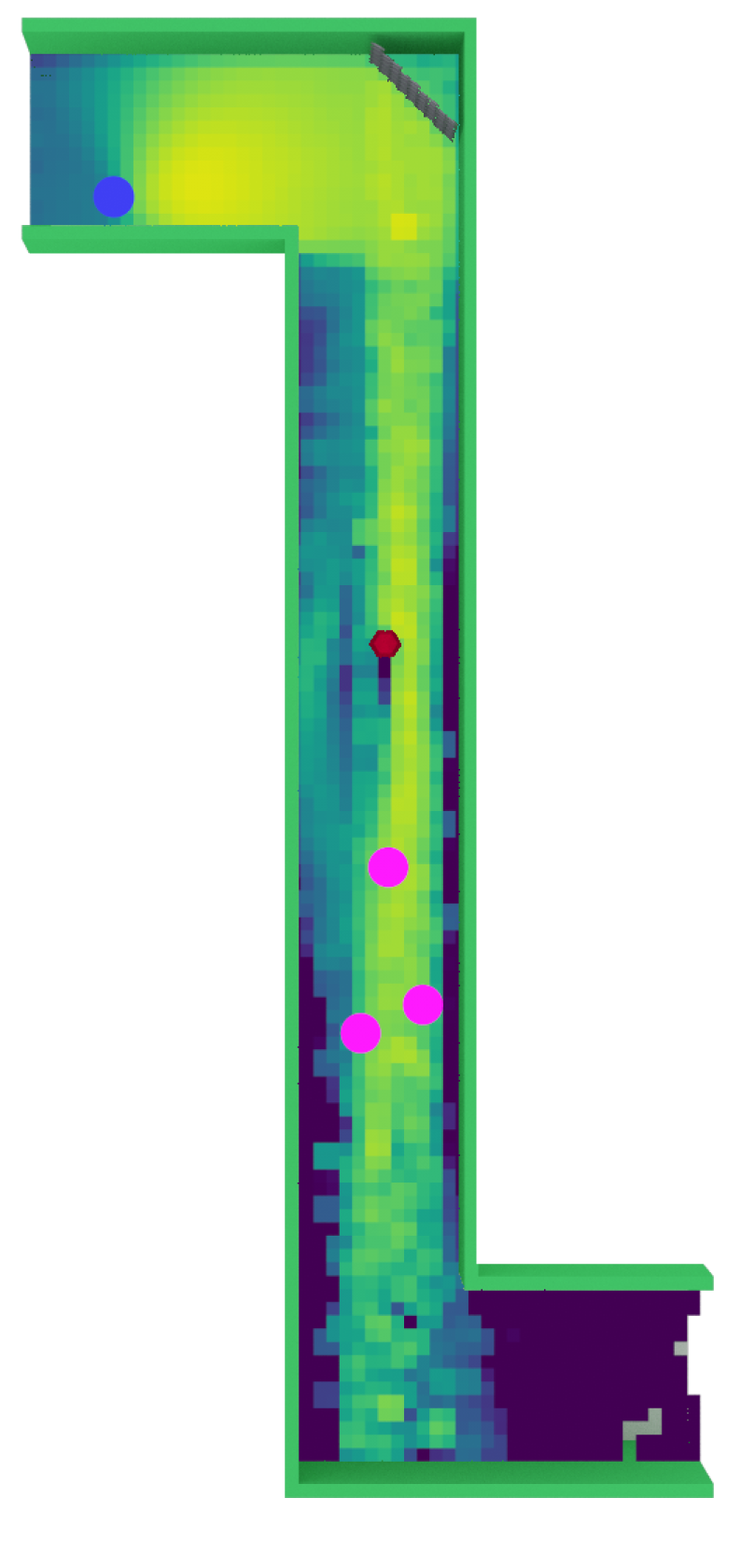}%
\label{Figure:beamfocusing_ma}}
\hfill
\subfloat{\includegraphics[width=0.36\linewidth]{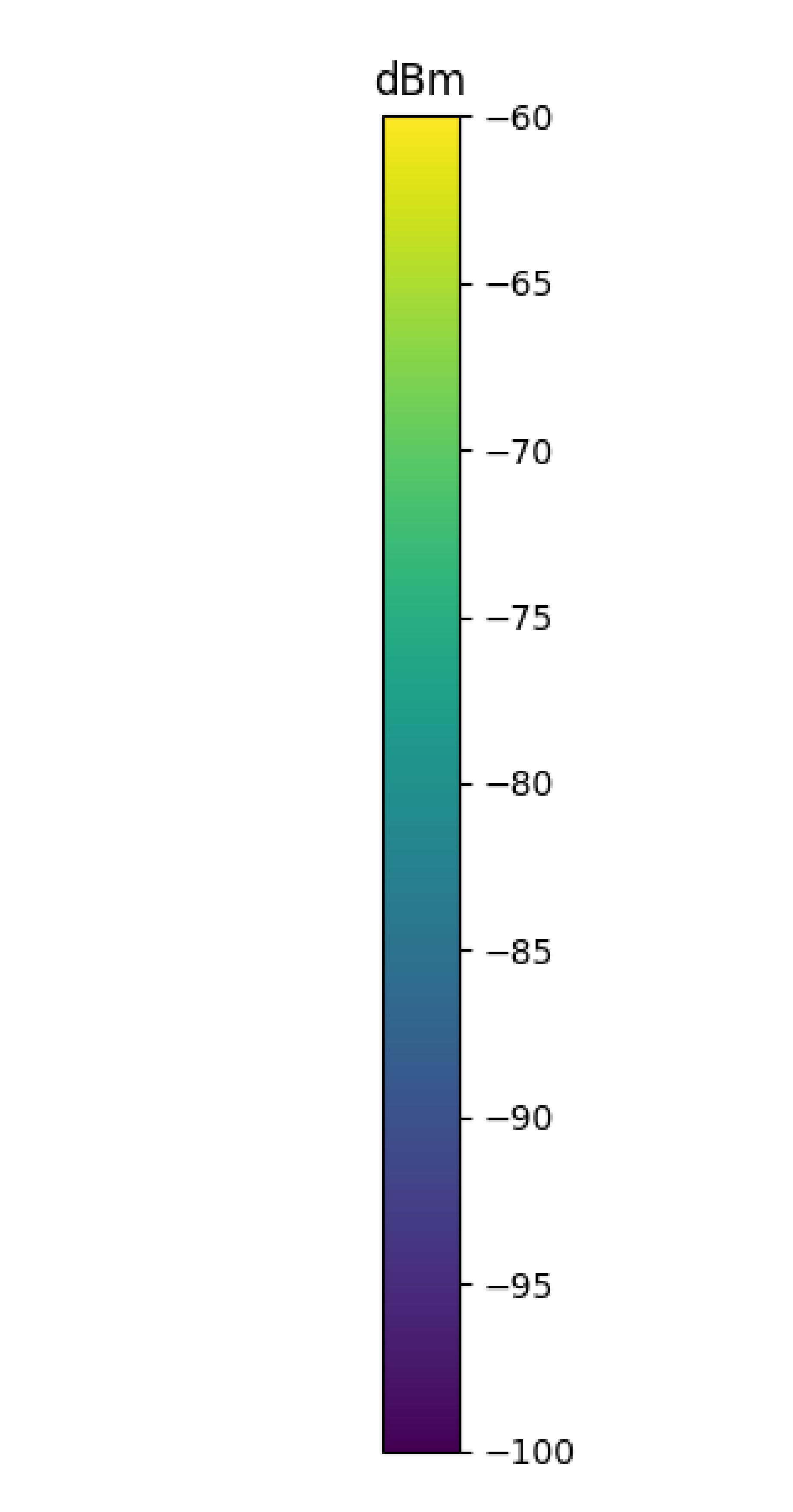}%
\label{Figure:scale}}

\caption{Heat map visualizations of spatial signal focusing capabilities: (a) an unassisted NLOS environment and (b) a conventional flat reflector. The proposed (e) multi-agent beam-focusing model demonstrates superior spatial selectivity compared to both the (c) single-agent and (d) column-based multi-agent baselines. RSSI values are color-coded in dBm; user positions are identified by purple circles, and the cylindrical obstacle is depicted in brown.}
\label{Figure:baseline_comparison}
\end{figure}

\begin{figure}[!ht]
    \centering
    \captionsetup{justification=centering}
    \includegraphics[width=1\linewidth]{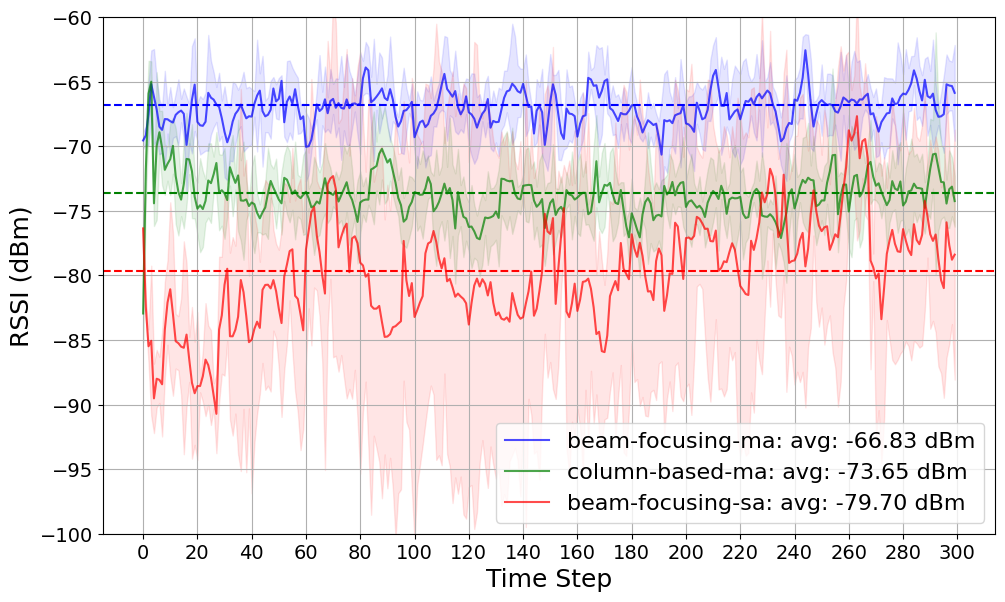}
    \caption{{Temporal RSSI performance under dynamic user mobility. Over a 300-step evaluation period, the proposed multi-agent beam-focusing framework (beam-focusing-ma) achieves superior stability and signal strength, averaging -66.83 dBm. This outperforms both the hardware-constrained column-based approach (-73.65 dBm) and the single-agent baseline (-79.70 dBm). Solid lines trace the mean RSSI, shaded regions illustrate ±1 standard deviation, and horizontal dashed lines indicate the overall temporal averages.}}
    \label{Figure:baseline_eval}
\end{figure}

\subsection{Spatial and Temporal Signal Enhancement}

To comprehensively evaluate the practical impact of the learned policies, we benchmark the spatial signal distribution and temporal adaptation capabilities of our framework under dynamic user mobility. Users are subjected to stochastic movement patterns, with their coordinates updating every four simulation steps to rigorously test the system's real-time adaptability.

\subsubsection{Spatial RSSI Distribution}

The spatial heatmaps (Fig.~\ref{Figure:baseline_comparison}) visually demonstrate the profound advantage of intelligent multi-agent control. In the baseline scenario with no reflector, severe non-line-of-sight (NLOS) blockage degrades the average RSSI to a value of \(-110.50\,\mathrm{dBm}\). Introducing a static, flat reflector provides a fundamental but limited improvement, elevating the RSSI to \(-94.40\,\mathrm{dBm}\).

In contrast, the proposed multi-agent beam-focusing approach (beam-focusing-ma) dynamically synthesizes highly concentrated signal pockets directly over the user locations, achieving an optimal average RSSI of \(-67.54\,\mathrm{dBm}\). This represents a \(42.96\,\mathrm{dB}\) enhancement over the no-reflector baseline and a \(26.86\,\mathrm{dB}\) gain over the static flat reflector. The multi-agent approach also outperforms the single-agent baseline (\(-72.48\,\mathrm{dBm}\)) and the hardware-constrained column-based method (\(-74.79\,\mathrm{dBm}\)), demonstrating that decentralized control over the full spatial degrees of freedom is essential for maximizing spatial selectivity.

\subsubsection{Temporal Adaptation and Stability}

Fig.~\ref{Figure:baseline_eval} tracks the temporal changes of the average RSSI across a 300-step evaluation period characterized by frequent user location shifts. The beam-focusing-ma framework maintains stable signal coverage, achieving an average RSSI of \(-66.83\,\mathrm{dBm}\) over the entire duration. Crucially, the multi-agent approach exhibits rapid recovery dynamics following user movement events, typically requiring only a single simulation step to recalibrate the focal points and restore optimal beam alignment.

The hardware-constrained column-based-ma approach maintains a stable but lower average of \(-73.65\,\mathrm{dBm}\), explicitly quantifying the approximately \(6\,\mathrm{dB}\) performance cost incurred when simplifying the servo-motor hardware architecture (Fig.~\ref{Figure:baseline_eval}). Meanwhile, the single-agent baseline (beam-focusing-sa) struggles to adapt to the dynamic environment, averaging only \(-79.70\,\mathrm{dBm}\). The pronounced variance and lower average of the single-agent approach further validate the necessity of the multi-agent task decomposition for real-time mobile tracking.

\begin{figure}[!t]
    \centering
    \captionsetup{justification=centering}
    \includegraphics[width=1\linewidth]{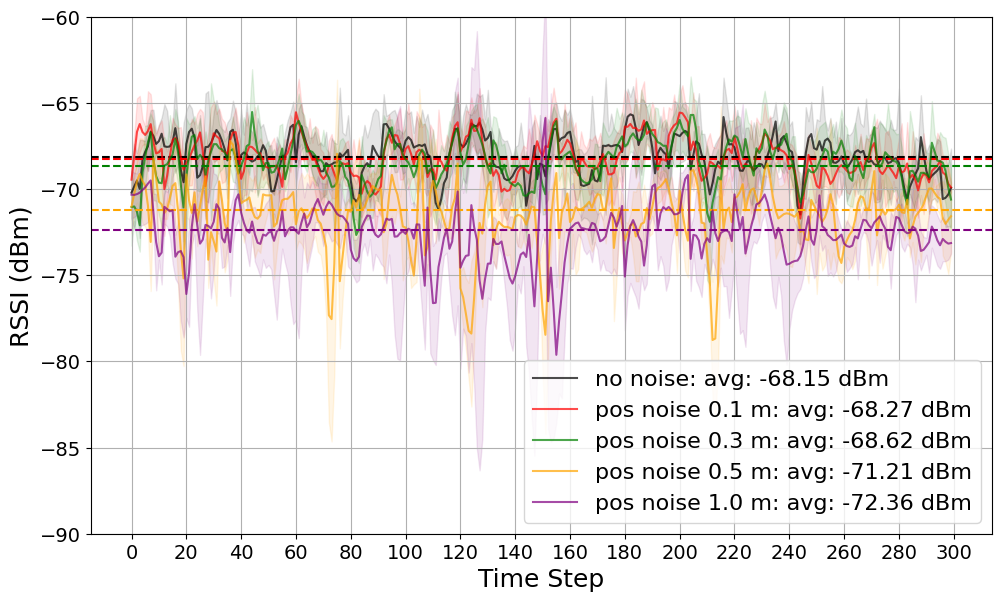}
    \caption{{Temporal RSSI performance under varying degrees of user localization noise. The system demonstrates graceful degradation, maintaining robust signal averages ranging from -68.15 dBm (ideal location data) to -72.36 dBm (1.0 m noise). Mean values are depicted by solid lines and overall temporal averages by dashed lines. The shaded regions (±1 standard deviation) illustrate the performance variance, visually highlighting the increased temporal instability the agents experience under severe positional uncertainty.}}
    \label{Figure:pos_noise_eval}
\end{figure}

\subsection{Robustness to Positioning Uncertainty}
Practical deployment scenarios inevitably involve imperfect user localization due to measurement uncertainties, sensor limitations, and varying environmental factors. To evaluate the resilience of the proposed multi-agent framework under deployment conditions, we introduce Gaussian noise with zero mean to the spatial coordinates of the users. This noise is added independently to each axis during both evaluation phases, utilizing standard deviations of 0.1, 0.3, 0.5, and 1.0 meters to simulate different tiers of indoor positioning accuracy.

The temporal RSSI performance evaluation (Fig.~\ref{Figure:pos_noise_eval}) reveals the system's robust adaptability to imperfect location data. The ideal scenario, relying on perfect location information, achieves an optimal average RSSI of \(-68.15\,\mathrm{dBm}\). Introducing positional noise results in a strictly graceful performance degradation rather than a catastrophic system failure. Under minor to moderate localization uncertainties, the framework maintains highly stable coverage, yielding average RSSI values of \(-68.27\,\mathrm{dBm}\) for \(0.1\,\mathrm{m}\) noise and \(-68.62\,\mathrm{dBm}\) for \(0.3\,\mathrm{m}\) noise. Most notably, even under severe positional uncertainty featuring a \(1.0\,\mathrm{m}\) standard deviation, the multi-agent system reliably sustains an average RSSI of \(-72.36\,\mathrm{dBm}\) throughout the evaluation period. Furthermore, while the average RSSI degrades gracefully, analysis of the performance variance (shaded regions in Fig.~\ref{Figure:pos_noise_eval}) reveals that severe positional noise (1.0 m) induces high temporal instability. The increased variance indicates that the agents must continuously readjust their focal points to target the true user location, highlighting that while the system remains operational, signal consistency is directly tied to localization accuracy.

This observed robustness stems directly from two fundamental design characteristics of our framework. First, the focal point control abstraction operates at a higher spatial level than individual tile control, which inherently reduces the system's sensitivity to precise positional accuracy requirements. The macroscopic geometric relationship between the focal points and user locations remains approximately valid despite moderate positioning errors. Second, the multi-agent coordination provides substantial spatial diversity across the reflecting elements, enabling the system to compensate for localization uncertainties through distributed optimization.

\section{Conclusion}
\label{sec:conclusion}

This paper presented a MARL framework for controlling mechanically adjustable metallic reflector arrays, effectively bypassing the CSI estimation bottleneck of conventional RIS. By utilizing a spatial focal point control abstraction within a CTDE paradigm, the system autonomously optimizes signal propagation based solely on user location. High-fidelity ray-tracing simulations demonstrated that this multi-agent beam-focusing approach yields up to a 26.86 dB improvement over static flat reflectors in dynamic non-line-of-sight environments. Moreover, the framework demonstrated vastly superior convergence, spatial signal concentration, and temporal stability when compared to single-agent and restricted multi-agent baseline architectures. The system also maintains robust signal coverage even under severe 1.0 m user positioning noise. Future work will focus on experimental validation through physical hardware prototypes and the integration of these MARL policies with commercial indoor positioning systems to realize fully adaptive smart radio environments.

\bibliographystyle{IEEEtran}
\bibliography{IEEEabrv,./main_ref}

\begin{thebibliography}{10}
\providecommand{\url}[1]{#1}
\csname url@samestyle\endcsname
\providecommand{\newblock}{\relax}
\providecommand{\bibinfo}[2]{#2}
\providecommand{\BIBentrySTDinterwordspacing}{\spaceskip=0pt\relax}
\providecommand{\BIBentryALTinterwordstretchfactor}{4}
\providecommand{\BIBentryALTinterwordspacing}{\spaceskip=\fontdimen2\font plus
\BIBentryALTinterwordstretchfactor\fontdimen3\font minus \fontdimen4\font\relax}
\providecommand{\BIBforeignlanguage}[2]{{%
\expandafter\ifx\csname l@#1\endcsname\relax
\typeout{** WARNING: IEEEtran.bst: No hyphenation pattern has been}%
\typeout{** loaded for the language `#1'. Using the pattern for}%
\typeout{** the default language instead.}%
\else
\language=\csname l@#1\endcsname
\fi
#2}}
\providecommand{\BIBdecl}{\relax}
\BIBdecl

\bibitem{direnzo:2020}
M.~Di~Renzo, A.~Zappone, M.~Debbah, M.-S. Alouini, C.~Yuen, J.~de~Rosny, and S.~Tretyakov, ``{Smart Radio Environments Empowered by Reconfigurable Intelligent Surfaces: How It Works, State of Research, and The Road Ahead},'' \emph{{IEEE Journal on Selected Areas in Communications}}, vol.~38, no.~11, pp. 2450--2525, 2020.

\bibitem{bjornson2022reconfigurable}
E.~Björnson, H.~Wymeersch, B.~Matthiesen, P.~Popovski, L.~Sanguinetti, and E.~de~Carvalho, ``{Reconfigurable Intelligent Surfaces: A Signal Processing Perspective with Wireless Applications},'' \emph{{IEEE Signal Processing Magazine}}, vol.~39, no.~2, pp. 135--158, 2022.

\bibitem{pan2022overview}
C.~Pan, G.~Zhou, K.~Zhi, S.~Hong, T.~Wu, Y.~Pan, H.~Ren, M.~D. Renzo, A.~Lee~Swindlehurst, R.~Zhang, and A.~Y. Zhang, ``{An Overview of Signal Processing Techniques for RIS/IRS-Aided Wireless Systems},'' \emph{{IEEE Journal of Selected Topics in Signal Processing}}, vol.~16, no.~5, pp. 883--917, 2022.

\bibitem{kim2022practical}
S.~Kim, H.~Lee, J.~Cha, S.-J. Kim, J.~Park, and J.~Choi, ``{Practical Channel Estimation and Phase Shift Design for Intelligent Reflecting Surface Empowered MIMO Systems},'' \emph{{IEEE Transactions on Wireless Communications}}, vol.~21, no.~8, pp. 6226--6241, 2022.

\bibitem{a9400843}
C.~Hu, L.~Dai, S.~Han, and X.~Wang, ``{Two-Timescale Channel Estimation for Reconfigurable Intelligent Surface Aided Wireless Communications},'' \emph{{IEEE Transactions on Communications}}, vol.~69, no.~11, pp. 7736--7747, 2021.

\bibitem{huang2020reconfigurable}
C.~Huang, R.~Mo, and C.~Yuen, ``{Reconfigurable Intelligent Surface Assisted Multiuser MISO Systems Exploiting Deep Reinforcement Learning},'' \emph{{IEEE Journal on Selected Areas in Communications}}, vol.~38, no.~8, pp. 1839--1850, 2020.

\bibitem{choi2024deep}
H.~Choi, L.~V. Nguyen, J.~Choi, and A.~L. Swindlehurst, ``{A Deep Reinforcement Learning Approach for Autonomous Reconfigurable Intelligent Surfaces},'' in \emph{{2024 IEEE International Conference on Communications Workshops (ICC Workshops)}}, 2024, pp. 208--213.

\bibitem{sheen2021deep}
B.~Sheen, J.~Yang, X.~Feng, and M.~M.~U. Chowdhury, ``{A Deep Learning Based Modeling of Reconfigurable Intelligent Surface Assisted Wireless Communications for Phase Shift Configuration},'' \emph{{IEEE Open Journal of the Communications Society}}, vol.~2, pp. 262--272, 2021.

\bibitem{aa11322690}
H.~Le, O.~Bedir, M.~Ibrahim, J.~Tao, and S.~Ekin, ``{Signal Whisperers: Enhancing Wireless Reception Using DRL-Guided Reflector Arrays},'' \emph{IEEE Transactions on Machine Learning in Communications and Networking}, vol.~4, pp. 265--281, 2026.

\bibitem{a8972365}
W.~Khawaja, O.~Ozdemir, Y.~Yapici, F.~Erden, and I.~Guvenc, ``{Coverage Enhancement for NLOS mmWave Links Using Passive Reflectors},'' \emph{{IEEE Open Journal of the Communications Society}}, vol.~1, pp. 263--281, 2020.

\bibitem{le2024guiding}
H.~Le, O.~Bedir, M.~Ibrahim, J.~Tao, and S.~Ekin, ``{Guiding Wireless Signals with Arrays of Metallic Linear Fresnel Reflectors: A Low-cost, Frequency-versatile, and Practical Approach},'' in \emph{{2024 IEEE 100th Vehicular Technology Conference (VTC2024-Fall)}}, 2024, pp. 1--7.

\bibitem{busoniu2008comprehensive}
L.~Busoniu, R.~Babuska, and B.~De~Schutter, ``{A Comprehensive Survey of Mmultiagent Reinforcement Learning},'' \emph{{IEEE Transactions on Systems, Man, and Cybernetics, Part C (Applications and Reviews)}}, vol.~38, no.~2, pp. 156--172, 2008.

\bibitem{zhang2018fully}
K.~Zhang, Z.~Yang, H.~Liu, T.~Zhang, and T.~Basar, ``{Fully Decentralized Multi-agent Reinforcement Learning with Networked Agents},'' in \emph{{International Conference on Machine Learning}}.\hskip 1em plus 0.5em minus 0.4em\relax {PMLR}, 2018, pp. 5872--5881.

\bibitem{a10261304}
B.~Hazarika, K.~Singh, S.~Biswas, S.~Mumtaz, and C.-P. Li, ``{Multi-Agent DRL-Based Task Offloading in Multiple RIS-Aided IoV Networks},'' \emph{{IEEE Transactions on Vehicular Technology}}, vol.~73, no.~1, pp. 1175--1190, 2024.

\bibitem{a10654286}
K.~Qi, Q.~Wu, P.~Fan, N.~Cheng, Q.~Fan, and J.~Wang, ``{Reconfigurable Intelligent Surface Assisted VEC Based on Multi-Agent Reinforcement Learning},'' \emph{{IEEE Communications Letters}}, vol.~28, no.~10, pp. 2427--2431, 2024.

\bibitem{nasari2022benchmarking}
\BIBentryALTinterwordspacing
A.~Nasari, H.~Le, R.~Lawrence, Z.~He, X.~Yang, M.~Krell, A.~Tsyplikhin, M.~Tatineni, T.~Cockerill, L.~Perez, D.~Chakravorty, and H.~Liu, ``{Benchmarking the Performance of Accelerators on National Cyberinfrastructure Resources for Artificial Intelligence / Machine Learning Workloads},'' in \emph{{Practice and Experience in Advanced Research Computing 2022: Revolutionary: Computing, Connections, You}}, ser. PEARC '22.\hskip 1em plus 0.5em minus 0.4em\relax New York, NY, USA: Association for Computing Machinery, 2022. [Online]. Available: \url{https://doi.org/10.1145/3491418.3530772}
\BIBentrySTDinterwordspacing

\bibitem{le2024insight}
\BIBentryALTinterwordspacing
H.~Le, Z.~He, M.~Le, D.~Chakravorty, L.~M. Perez, A.~Chilumuru, Y.~Yao, and J.~Chen, ``{Insight Gained from Migrating a Machine Learning Model to Intelligence Processing Units},'' in \emph{{Practice and Experience in Advanced Research Computing 2024: Human Powered Computing}}, ser. PEARC '24.\hskip 1em plus 0.5em minus 0.4em\relax New York, NY, USA: Association for Computing Machinery, 2024. [Online]. Available: \url{https://doi.org/10.1145/3626203.3670527}
\BIBentrySTDinterwordspacing

\bibitem{qualcomm2024unlocking}
\BIBentryALTinterwordspacing
{Qualcomm}. (2024) {Unlocking On-device Generative AI with an NPU and Heterogeneous Computing}. [Online]. Available: \url{https://www.qualcomm.com/content/dam/qcomm-martech/dm-assets/documents/Unlocking-on-device-generative-AI-with-an-NPU-and-heterogeneous-computing.pdf}
\BIBentrySTDinterwordspacing

\bibitem{yu2022surprising}
C.~Yu, A.~Velu, E.~Vinitsky, J.~Gao, Y.~Wang, A.~Bayen, and Y.~Wu, ``{The Surprising Effectiveness of PPO in Cooperative Multi-agent Games},'' \emph{{Advances in Neural Information Processing Systems}}, vol.~35, pp. 24\,611--24\,624, 2022.

\end{thebibliography}

\end{document}